# Developing a general-purpose clinical language inference model from a large corpus of clinical notes

Madhumita Sushil, PhD[1,*]

Dana Ludwig, MD[1]

Atul J. Butte, MD, PhD[1,2,3†]

Vivek A. Rudrapatna, MD, PhD[1,4,†]

1. Bakar Computational Health Sciences Institute, University of California, San Francisco, San Francisco, CA, USA
2. Center for Data-driven Insights and Innovation, University of California, Office of the President, Oakland, CA, USA
3. Department of Pediatrics, University of California, San Francisco, CA, 94158, USA
4. Division of Gastroenterology, Department of Medicine, University of California, San Francisco, San Francisco, CA, USA
Emails: Firstname.Lastname@ucsf.edu

[*]Corresponding Author
Bakar Computational Health Sciences Institute, 490 Illinois Street, Cubicle 2219, 2nd Fl, North Tower, San Francisco, CA 94143
Telephone: +1 (415)-514-1971

[†]Equal Contribution

**Word count**

3989 words excluding the structured abstract, tables, figures, footnotes, references, and acknowledgments.

# ABSTRACT

**OBJECTIVE**

Several biomedical language models have already been developed for clinical language inference. However, these models typically utilize general vocabularies and are trained on relatively small clinical corpora. We sought to evaluate the impact of using a domain-specific vocabulary and a large clinical training corpus on the performance of these language models in clinical language inference.

**MATERIALS AND METHODS**

We trained a Bidirectional Encoder Decoder from Transformers (BERT) model using a diverse, deidentified corpus of 75 million deidentified clinical notes authored at the University of California, San Francisco (UCSF). We evaluated this model on several clinical language inference benchmark tasks: clinical and temporal concept recognition, relation extraction and medical language inference. We also evaluated our model on two tasks using discharge summaries from UCSF: diagnostic code assignment and therapeutic class inference.

**RESULTS**

Our model performs at par with the best publicly available biomedical language models of comparable sizes on the public benchmark tasks, and is significantly better than these models in a within-system evaluation on the two tasks using UCSF data. The use of in-domain vocabulary appears to improve the encoding of longer documents.

**DISCUSSION**

The use of large clinical corpora appears to enhance document encoding and inferential accuracy. However, further research is needed to improve abbreviation resolution, and numerical, temporal, and implicitly causal inference.

**CONCLUSION**

We have developed a clinical language model using a large deidentified clinical corpus and find that in-domain training data appears to be more important than diverse but less relevant training corpora.

**OBJECTIVE**

Clinical notes recorded by practitioners in Electronic Health Records (EHRs) have a distinct vocabulary and writing style compared to other publicly available corpora like Wikipedia, textbooks, scientific articles, and social media data found on the internet. These textual reports often assume knowledge of the specialty they are written for and are replete with domain-specific abbreviations and terminology[1]. Additionally, clinical notes can contain both semi-structured and unstructured, ungrammatical text[2]. These text reports are frequently copy-forwarded from one patient encounter to the next[3,4]. Furthermore, new notes are added for every encounter across most clinical departments, which makes the clinical rhetoric dense with several temporal references across the longitudinal history. These features make clinical language inference uniquely challenging and suggest the potential need for dedicated language models that have been specifically trained to draw appropriate inferences from clinical text.

Here, we report the development of a new clinical language model trained on a large corpus of notes documented at a large academic medical center — the University of California, San Francisco (UCSF) Health. Key features of our approach include the incorporation of large domain-specific terminology and language structure from clinical documentation. By incorporating a wide range of clinical notes from different specialties, written for different purposes by varied provider types, we expose our model to a diverse set of clinical topics, concepts, and vocabulary during training. We aim to improve transfer learning for clinical language inference applications by developing a language model that has learned from a diverse, domain-specific dataset. We evaluate our model on a wide range of well-known clinical reasoning challenges that are directly pertinent to many real-world clinical research tasks: identifying clinical concepts of type problems, treatments, tests, and inferring relations between them from clinical text, identifying the mentions of temporal expressions, clinical departments, evidentials, and occurrences, inferring whether pairs of sentences are semantically equivalent, contradicting or neutral to each other, assigning diagnostic codes to a patient's discharge report, and inferring the therapeutic classes of medications administered to a patient from their discharge reports. We find that a language model trained solely on text specific to a large medical center performs well on all these benchmarks and seems to be useful for a broad range of downstream language-related tasks as conducted within a given medical center. The promising performance of the model also highlights the utility of developing large, deidentified EHR datasets at medical centers to support clinical research and improve the state-of-the-art performance in clinical language processing.

**BACKGROUND AND SIGNIFICANCE**

Language modeling has a long history in NLP and different statistical techniques have been historically used to learn the semantics of sequences of tokens in the text[5]. Bidirectional Encoder Representations from Transformers (BERT) models[6] have recently become popular for language modeling due to their ability to learn these semantics efficiently from large, unannotated corpora by using contextual and distributed representations of token sequences in self-supervised setups. These models have been used extensively for transfer learning on several popular NLP tasks, where they have held state-of-the-art performance[7][1]. A suite of similar language models based on the transformers architectures, albeit using modified optimization schemes, have since followed[8–14].

These models have also been adopted in the biomedical domain, where biomedical data including abstracts and articles (through PubMed, PubMed Central, and Semantic Scholar), and the publicly available MIMIC-III clinical corpus[15,16] are used for developing large language models[17–26]. These models have shown superior performance compared to those trained on other general-purpose corpora like Wikipedia, BookCorpus, and CommonCrawl data. However, the limitations of these models include the limited size of publicly available medical data, resulting in reduced exposure to the range of vocabulary and language used across different medical specialties. In this study, we aim to explore the benefit of incorporating a large, deidentified clinical corpus for training these general-purpose clinical language models. We compare the performance of our model to different publicly available clinical language models and provide further discussions ahead.

---

[1] Please refer to https://super.gluebenchmark.com/leaderboard for the most recent leaderboard results.

To the best of our knowledge, the study most closely related to ours is the contemporary work of training the new GatorTron clinical language model[26]. This model is a Megatron model[13] trained on a large corpus of deidentified clinical notes from University of Florida network jointly with text from biomedical articles, the MIMIC-III corpus, and Wikipedia. The smallest GatorTron model is three times the size of our model and has been trained on a pod of 992 GPUs.

**MATERIALS AND METHODS**

**Training data**

Under the UCSF IRB (#21-34088), we have trained a domain-specific, transferable BERT baseline (base) language model on a corpus of 75 million clinical notes at UCSF Health, de-identified as previously described[27]. This corpus of 39 billion words encompasses data from 2.25 million patients spanning 28 million encounters between 2012 and early 2021. These notes record fine-grained details of patient visits and reflect the advanced medical knowledge of care providers at a tertiary care center in the United States. The corpus is significantly larger than many other publicly available clinical and biomedical corpora, as shown in Table 1. Some frequent note types in the corpus include imaging reports, progress notes, telephonic encounter notes, consults, emergency department notes, pathology and cytology reports, ECG reports, assessment and plan notes, procedure-related notes, and discharge summaries. This large diversity in categories of notes also reflects that the corpus is more heterogeneous than publicly available clinical corpora. The complete list of the categories of notes in the training corpus and their frequencies can be referred to in the Supplementary Materials, Table S1.

**Table 1: Comparison between the size of the UCSF notes corpus, as measured by word count, with publicly available biomedical and clinical corpora used to train transformers-based language models**

| Corpus | UCSF notes | PMC full-text articles | PubMed abstracts | MIMIC-III notes | Scientific Papers (SciBERT) | English Wikipedia | Books Corpus |
|---|---|---|---|---|---|---|---|
| **Size** | 39.1B | 13.5B | 4.5B | 0.5B | 3.2B | 2.5B | 0.8B |

**Training scheme**

*Tokenization and vocabulary*

Vocabulary size for subword tokenizers is a predetermined hyperparameter that corresponds to the upper bound on the number of tokens created from a corpus. After the vocabulary is generated, input words that do not map to the model vocabulary are split into subword tokens. As the vocabulary size increases, so do the number of parameters in language models, since the embedding layer is dependent on it. To account for the large size and diversity of our corpus, we have incorporated a larger custom clinical vocabulary of 64,000 cased tokens by training a BERT wordpiece tokenizer[28] on the UCSF corpus[2], instead of 35,000–50,000 tokens that are frequently used. We hypothesized that a larger clinical vocabulary would make the model more suitable for processing long clinical documents by encoding domain-specific context more concisely.

---

[2] Punctuation symbols are space-separated from the nearby cased alphabets before training the tokenizer.

*Optimization specifics for model pretraining*

Our model training was distributed across four 20GB NVIDIA GPUs within a single node of the Sun Grid Engine (SGE) high-performance computing cluster at UCSF. We additionally used Message Passing Interface (MPI) and Accelerated Linear Algebra (XLA) optimization via NVIDIA NGC singularity containers on this computing cluster. The training was completed in two phases with the Adam optimizer[29]: i) In the first phase, we trained the UCSF-BERT (base) model on sequence lengths of up to 128 tokens for 2/3rd of the training time of 500k steps. ii) In the second phase, we continued to train the same model, but with the maximum sequence length of 512 tokens, for the remaining nearly 1/3rd of the training time or 275k steps. This two-phased approach was critical to utilizing a large corpus in an academic setup. The complexity of training a transformers model is of the order $N^2$ in sequence length. Doing so in two phases speeds up the training process significantly, as the model converges to a good parameter distribution relatively faster when training on shorter sequences. In the first phase, the model efficiently learns the embeddings for the first 128 token positions and requires fewer steps for converging on longer sequences later. Across both phases, we iterated over our entire corpus nearly once, although we anticipate that training it for longer may further improve the performance on downstream tasks.

To approximate the gradients more accurately before backpropagation and reduce the number of steps required for convergence, we use gradient accumulation and incorporate an effective batch size of 2048 examples during training (32 batches * 64 examples).

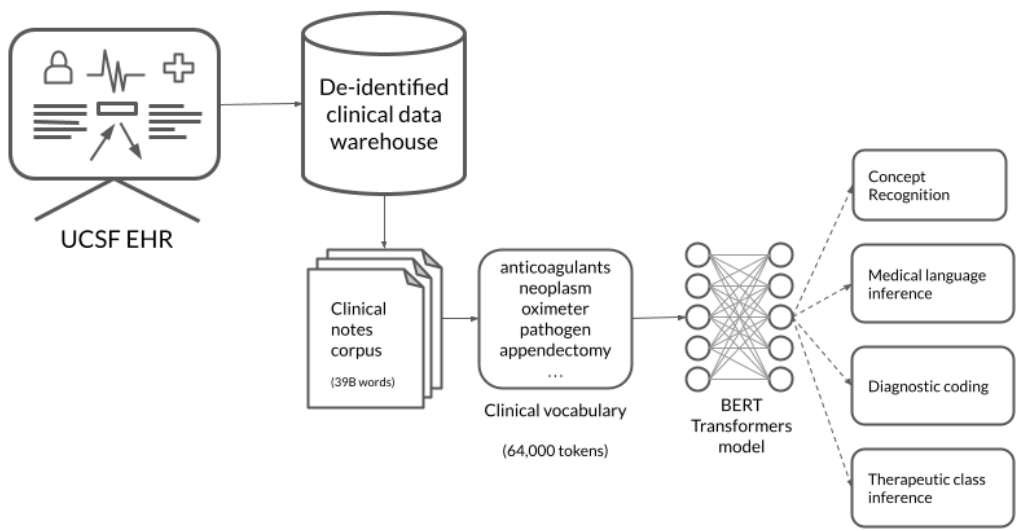

**Figure 1:** Schematic diagram of the UCSF-BERT training and evaluation pipeline

**Evaluation**

*Public benchmarks*

In line with the prior literature[22], we have benchmarked our model on three popular publicly available clinical language processing datasets: 2010 i2b2 shared tasks for concept recognition and relation classification[30], 2012 i2b2 shared task for temporal entity extraction[31], and medical language inference (MedNLI)[32]. These tasks are described in further detail in Supplementary Materials, Section S-A. We have not evaluated the performance on prior biomedical tasks, such as those related to named entity recognition of chemicals, proteins, and genes. We instead focus exclusively on performing evaluations on clinical tasks that can be inferred from information in Electronic Health Records (EHRs). We have also excluded evaluation on the i2b2-2014 Protected Health Information (PHI) identification task[33] due to the inherent differences in the notation of PHI between the UCSF dataset and the MIMIC-III critical care dataset[15,16] that these tasks are based on.

*UCSF-specific tasks*

We have additionally evaluated the UCSF-BERT model on two clinical tasks on the UCSF data. These tasks consist of discharge reports for patient encounters at UCSF and labels related to the encounter. The text reports for these tasks are longer than the public benchmarks we have evaluated and provide a within-domain evaluation of the model's performance. For both tasks, we only retain discharge summaries longer than 2000 characters in length and exclude the discharge summaries recorded by nursing providers. If multiple discharge reports are found for the same encounter, we retain the longest report as the discharge summary pertaining to that encounter.

1) **Top 50 ICD coding:** In the first task, we implemented the popular setup of assigning International Classification of Diseases (ICD) diagnostic codes to discharge summaries from the list of 50 most frequent ICD codes in the dataset[34]. This is a multi-label classification task because every encounter can be assigned multiple ICD codes corresponding to multiple diagnoses. To reduce the complexity of mapping ICD-9 and ICD-10 codes, we restrict ourselves only to those encounters that contain a valid ICD-9 code and exclude ICD-10 codes from the current analysis. The complete list of the most frequent ICD-9 codes in our dataset can be found in Table S2 in Supplementary Materials. This task setup contains 28k discharge reports at UCSF. They have further been split into 8:1:1 train:dev:test sets by patient IDs such that the same patients do not repeat between different subsets.
2) **Therapeutic class prediction:** In this task, we classify the therapeutic classes of the medications ordered for or administered to a patient during an encounter. These therapeutic classes are inferred from the ingredients present in medications and are based on Anatomic Therapeutic Classification (ATC)[35]. These classes are already present as structured data in our EHR, and fall into one of the 50 broad categories of medications such as *vitamins*, *antibiotics*, *antineoplastics*, etc. The complete list of these therapeutic classes can be found in Table S3 in Supplementary Materials. Given that every patient could be administered multiple medications during a stay, this task is also a multi-label classification task. For this task, we obtain a dataset of 22k discharge reports at UCSF, which we have split in the same manner as the ICD-coding dataset.

Due to its focus on document-level classification, the UCSF-specific tasks contain instances that are an order of magnitude longer than the public benchmark tasks (all of which are sentence-level). The median length of UCSF discharge reports is at least 1k tokens, whereas public tasks contain a median instance length of at most 28 tokens. These numbers can be referred to in further detail in Table S4 in the Supplementary Materials.

We compare the performance of our UCSF-BERT model with several popular publicly-available biomedical and clinical language models: the BioBERT model[21], the clinical BERT model[17], the SciBERT model[18], the Biomed-RoBERTa model[20], the BioLM model[22], and the GatorTron-OG model[26] (See Section S-B in the Supplementary Materials for details). To ensure a fair comparison, we have only included the results from models of comparable sizes (in terms of the number of parameters), i.e, baseline (base) transformers models. In addition, for comparison, we evaluated the much larger GatorTron-OG model to illustrate the impact of a greatly increased parameter count (345M vs 135M) on the model performance.

*Effect of sequence length*

To analyze the impact of sequence length, we developed a second version of the UCSF-BERT model that supports the maximum sequence length of only 128 tokens. After an initial 500k steps of model training, we continue to train this model for an additional 275k steps to stop at the same iteration as the UCSF-BERT (512) model. We compare the performance of this 128-length model with the 512-length model to analyze the impact of sequence length on downstream tasks. Both models are trained for the same number of optimization steps. Hence, any differences in the obtained results would primarily be due to the differences in the processing limits of the two UCSF-BERT models.

**RESULTS**

In Table 2 and Figure 2, we present the results of named entity recognition and classification tasks. These results are computed as the median over 5 runs with different seeds. The confidence interval of the performance is indicated as the standard deviation over these 5 runs and can be analyzed in Figure 2.

**Table 2**: Performance of biomedical and clinical transformers models (base sizes)[3]. All scores are median over 5 runs. The bolded values represent the best performance, and the underlined values refer to the results within 0.5% of the best model. We do not report the parameters for the UCSF-BERT (128) because we use the same architecture as UCSF-BERT (512), albeit in a limited sequence length setting.

| Model / Task | Metric | BioBERT | Clinical BERT | SciBERT | BioMed RoBERTa | BioLM RoBERTa | UCSF-BERT (512) | UCSF-BERT (128) | Gator-Tron OG |
|---|---|---|---|---|---|---|---|---|---|
| # of parameters | N/A | 110M | 110M | 110M | 125M | 125M | 135M | N/A | 345M |
| i2b2 2010 NER | %F1 | 86.0 | 86.3 | 86.3 | 85.0 | <u>88.1</u> | **88.3** | <u>88.2</u> | **89.1** |
| i2b2 2012 NER | %F1 | 77.6 | 78.0 | 77.6 | 76.4 | **79.5** | <u>79.3</u> | <u>79.1</u> | **80.2** |
| i2b2 2010 RE | %Micro F1 | 74.4 | 74.0 | 69.8 | 75.0 | 75.0 | **75.7** | 72.6 | **77.4** |
| MedNLI | %Accuracy | 82.5 | 81.8 | 79.7 | 85.1 | **87.1** | <u>86.8</u> | 84.8 | **88.6** |
| ICD9-top50 coding | %Micro F1 | 52.0 | 51.0 | 53.7 | 52.5 | 54.4 | **56.2** | 39.1 | **56.7** |
| Therapeutic class prediction | %Micro F1 | 74.3 | 74.0 | 74.8 | 75.0 | 76.0 | **77.3** | 72.5 | **77.4** |

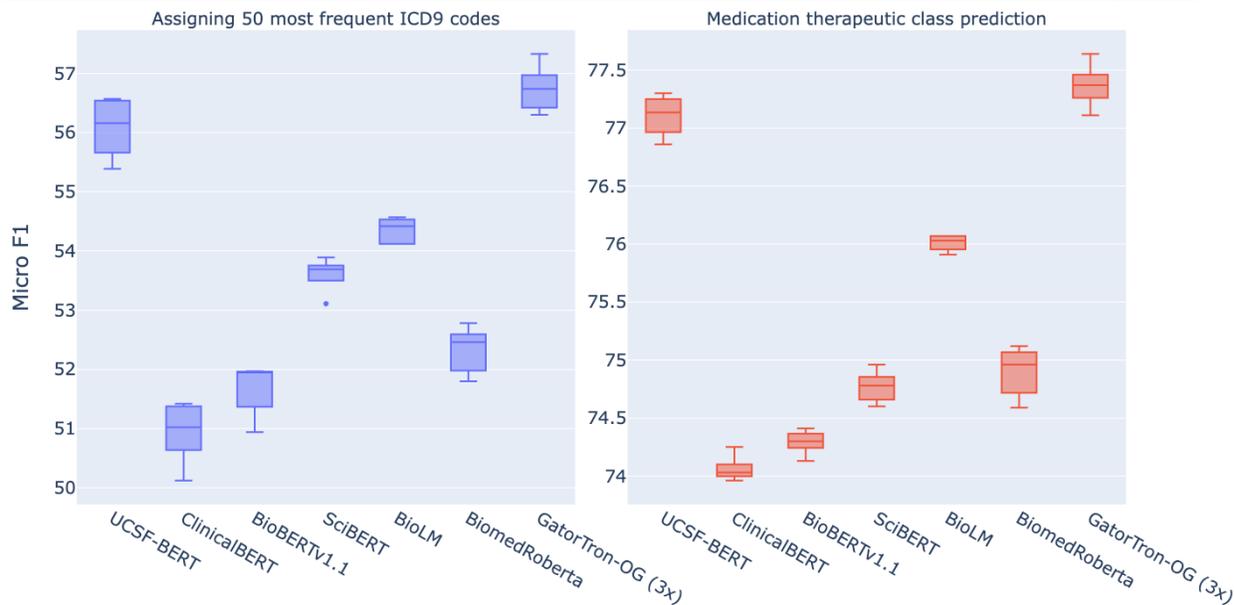

**Figure 2:** Performance of publicly available base biomedical and clinical transformers models on the ICD9-top50

---

[3] Note that the number of parameters in the smallest Gatortron (base) model is still thrice the other models. We have compared our performance to this model nonetheless since it uses the largest clinical corpus for pretraining.

coding and the therapeutic class prediction tasks on UCSF data. The confidence intervals are computed from results over 5 runs with different seeds. The UCSF-BERT model here supports a sequence length of up to 512 tokens.

We find that the UCSF-BERT (512) model performs at par with public models on all the public benchmark tasks and holds state-of-the-art performance on the internal UCSF tasks. This highlights the utility of this model for clinical language inference across a wide range of fundamental NLP tasks on EHR data. Moreover, the multi-label task of classifying the correct ICD-9 codes from discharge summaries is significantly more challenging than classifying the therapeutic classes for prescribed medications from these notes.

Furthermore, training language models on the same data source shows a significant benefit. All the publicly available benchmark tasks are based on text from the MIMIC-III corpus[15,16]. The same corpus is also included in the training data of the BioLM model and the GatorTron-OG model, which hold the best performance on these tasks[4]. The UCSF-BERT model that has been trained on clinical notes from the same institution performs the best on the internal evaluation tasks, comparable to the three-times larger GatorTron-OG model. Note that despite the seemingly larger median performance of the GatorTron-OG model, across multiple runs, it performed similarly to the UCSF-BERT model on these tasks (Figure 2).

We additionally find that the performance of the UCSF-BERT model that uses only 128 tokens is lower than that of the model that uses 512 tokens. The differences are higher for tasks containing longer instances in the dataset, with the difference being the most significant (>17% micro-F1) in inferring diagnostic codes from long discharge summaries. Only marginal differences are observed on the i2b2 2010 and 2012 NER tasks, which contain the shortest sequences of text.

**Impact of a custom vocabulary**

**Table 3**: **Vocabulary difference between UCSF-BERT, BERT (Wikipedia + BooksCorpus vocabulary), and BioLM models (PubMed vocabulary)**

| + UCSF vocab - BERT vocab | - UCSF vocab + BERT vocab | + UCSF vocab + BERT vocab | + UCSF vocab - PubMed vocab | - UCSF vocab + PubMed vocab | + UCSF vocab + PubMed vocab |
|---|---|---|---|---|---|
| anticoagulants | Communists | magnetic | nasolaryngoscope | isoprenaline | analgesia |
| neoplasm | Holocaust | patient | Parathyroidectomy | anhydride | nasopharynx |
| oximeter | Rwanda | pneumonia | endocrinology | semiconductors | pathologists |
| pathogen | Wildlife | spider | linaclotide | Boston | lung |
| appendectomy | Cyrillic | cheese | diverticula | planktonic | gastroenteritis |

Our custom vocabulary contains a larger number of clinical terms such as *neoplasm* and *anticoagulants*, instead of terms like *communists* and *playback* that are present in the vocabulary of the BERT model trained on Wikipedia and BooksCorpus[6], which has also been reused by the BioBERT[21] and ClinicalBERT[17] models (Table 3). We additionally find that specific clinical procedures are better reflected in custom UCSF vocabulary than the PubMed vocabulary, although PubMed better covers biomedical terms for proteins and chemical compounds.

---

[4] The ClinicalBERT model also uses the MIMIC-III corpus for model pretraining but performs worse on public benchmarks than other models. We believe it may benefit from either a custom vocabulary or a longer training time.

To analyze the impact of this clinical vocabulary, we assess the difference in sequence lengths (in terms of the number of tokens) when different vocabularies are used to encode the same text. BERT and RoBERTa models both use tokenizers that split out-of-vocabulary terms into subword units. This circumvents the issue of out-of-vocabulary terms, but the tokenizer vocabulary itself becomes important in determining token sequences, and consequently the sequence length of examples. We compute the average sequence length in the dataset by encoding the examples in our evaluation datasets with different vocabularies of the baseline models. The length of an example is defined as the number of tokens in that example, dependent on the vocabulary in use. We present the mean and the median number of tokens in the datasets, when encoded with different vocabularies, in Table S5 in the Supplementary Materials. We find that the EHR-specific vocabulary created from the UCSF de-identified clinical notes corpus provides tokenized clinical datasets that are shorter by nearly 20% compared to the BERT vocabulary derived from Wikipedia and BooksCorpus. The difference in lengths is smaller, albeit still significant (nearly 5%), with the BioLM vocabulary created from PubMed data. Fewer tokens are split into subword units when using the UCSF-BERT vocabulary because it contains several long tokens specific to the clinical rhetoric, for example, *hepatosplenomegaly*, *atherosclerotic*, *Hypercholesterolemia*, *retroperitoneal*, *hydrochlorothiazide*, etc.

**DISCUSSION**

Our results highlight the UCSF-BERT model is useful for many clinical inference tasks. Significant performance gains are visible in a within-institution evaluation compared to most publicly available biomedical language models, which highlights the benefit of training language models on in-domain data. This suggests that the UCSF-BERT model can be improved further on publicly available benchmark tasks with further pretraining on publicly available data sources — biomedical abstracts and articles, and the MIMIC-III corpus.

One explanation why the ICD coding task is more complex than the therapeutic class prediction task is that ICD codes incorporate a complex and imbalanced hierarchy[36], and are also noisy[37]. However, a linear, lexical mapping exists between the medication names and their therapeutic classes, making this inference task more learnable.

Furthermore, the performance drop in the model that uses only 128 tokens long sequences, as compared to 512 tokens, reinforces our understanding that models supporting longer sequences are more suitable for real-world clinical use cases that often require document-level or multi-document inferences. We can benefit further by exploring strategies that support even longer sequences during model pretraining, such as the Longformer[10] or the BigBird models[14]. However, these models are also computationally more expensive to train, which often makes their training infeasible.

In addition, we find that encoding the datasets with an in-domain vocabulary allows us to process significantly longer contexts compared to using out-of-domain publicly available models and vocabularies. This strategy is particularly beneficial for circumventing the sequence length limits imposed by the computational complexity of transformers models and making inferences at the document-level for capturing longer distance dependencies.

Finally, since our language model is trained on the clinical notes from a single institution, they may contain biases specific to the institution such as the patient and disease distribution, EHR vendor, writing style, etc. The ability to include clinical notes from multiple centers, which is challenging due to privacy concerns, would allow us to train a more generalizable language model.

**Error analysis**

*Named entity recognition*

On analyzing the errors made by the UCSF-BERT model on the i2b2 2010 concept recognition task (examples presented in Table S6 the Supplementary Materials), we find that most of the errors occur because the modifiers are included as a part of the annotated concepts in the dataset. The model either excludes them, includes too many modifiers, or joins concepts across modifiers. Although these errors are not critical for most downstream applications, it highlights the influence of the annotation guidelines on the model outputs.

Additionally, we find that the model has difficulty in inferring terms that are highly domain-specific and contextual, such as abbreviations and acronyms, e.g. *-nt*, *NGTD*, "started on *sips*", "advanced to *clears*". Although training the UCSF-BERT model on a large clinical corpus reduces these errors, further research is needed to provide additional context to these terminologies, particularly in the setups that require cross-institution transfer.

Furthermore, we find that our concept detection model is highly contextual. If the context is sufficient to infer the type of the concept, it can be recognized as such even if the concept phrase itself is ambiguous (for example, the italicized phrase was recognized as treatment in "Baby was vigorous at birth and received *routine DR Johnson*"). This finding reinforces the contextual ability of the transformers models, especially when trained in the BIO setup.

Lastly, we also find that the model mixes up concepts whose type may be either too generic (e.g. *problems* or *findings* annotated as the concept "Problem") or ambiguous (e.g. *incision* that can either be a treatment or a finding annotated as problem, or *C. difficile toxin* that can either be a problem or a test). These concepts are sometimes inconsistently annotated in the corpus, which can make it challenging for the model to learn them. For example, *further care* is annotated as the concept "Treatment", but *conservative management* is not. Similarly, *NG tubes* have been annotated as "Treatment" although they are a method for delivering treatment instead of the treatment itself. However, we also find that the model sometimes misses identifying simpler concepts, for example, treatment-related concepts *anticoagulated, daily transfusions,* and *pain control*, and problem-related concepts like *a previous trauma*.

*Classification*

Limitations of the BioBERT model when making inferences in the clinical domain have been demonstrated earlier[38]. In a similar spirit, we have explored the limitations of the UCSF-BERT model fine-tuned on the MedNLI dataset by evaluating it on a range of manually curated NLI instances, which can be referred to in Table S7 in the Supplementary Materials.
1) <u>Numeric inference:</u> We created several instances to reflect inference relations between laboratory values and clinical conditions/diagnoses. Particularly, we analyzed the model's understanding of blood glucose levels, blood pressure, BMI, pulse rate, and hematocrit to infer whether they correspond to hyper/hypo-glycemia, hyper/hypo-tension/shock, obesity/overweight/healthy weight, tachycardia/bradycardia, and anemia respectively. We found that the UCSF-BERT model can reliably infer the clinical state for frequent lab tests (e.g., glucose measurements) and numeric results (e.g., 600, 30). However, the model is not reliable in inferring from relatively infrequent tests such as those about serum calcium and albumin levels. This is expected based on prior literature demonstrating limited numeric reasoning skills of language models[39]. This reiterates the benefit of training language models on large clinical corpora that mention numeric results, as well as emphasizes the need for integrating external knowledge sources to encode infrequent clinical evidence.

2) <u>Inference between clinical description and the patient's state:</u> We further curated several instances where the premise discusses clinical rhetoric describing a patient's state, and the hypothesis is the potential conclusion or differential diagnosis for the patient (for example relating lower abdomen pain or tenderness with appendicitis, blood culture findings with infectious diseases, and medications with corresponding diseases). We found that the UCSF-BERT model performs surprisingly well in these cases, potentially due to the correlations learned from the clinical notes. However, more complex inferences requiring an understanding of causation, for example, adverse drug events, need further research for reliable inference.

3) <u>Temporal inference:</u> We analyzed whether the UCSF-BERT model can make inferences over a sequence of events to infer the temporal ordering of different events from the clinical rhetoric. We, unfortunately, found that the UCSF-BERT model is not good at inferring temporal ordering without being explicitly trained for the task. The clinical text contains several references to the temporal sequence of events both before the clinical presentation and during treatment. To support better inferences over complex clinical timelines, self-supervised pre-training objectives that encode temporality would provide better inductive biases to language models to perform these inferences.

**CONCLUSIONS**

A transformers model trained on a large corpus of clinical notes from the same health system as the downstream applications performs better inference on downstream tasks than transferring knowledge from publicly available biomedical and clinical models of comparable sizes. Furthermore, these models can process longer contexts when trained with a custom, domain-specific vocabulary. The advantage of domain-specific data reduces when using significantly larger transformers models trained on similar, but not the same, data sources. The UCSF-BERT model is excellent at making contextual inferences from explicit mentions in the text However, clinical transformers models still struggle with abbreviation and acronym resolution, temporal sequence inference, implicit causal inference, and infrequent numeric inference. Future directions include further research on long sequence encoding, temporal and numerical inference bolstering, and out-of-domain generalization. Incorporating diverse corpora during model training — biomedical abstracts and articles, the MIMIC-III corpus, as well as clinical notes from multiple health systems — would potentially improve the generalizability of these models.

**ACKNOWLEDGEMENTS**

We would like to acknowledge the Information Commons platform at UCSF that has provided the deidentified clinical dataset for our research, and the Wynton computing environment at UCSF that was used to train the model in a high-performance, distributed computing setup. Furthermore, we would like to thank several members of the Butte lab and the Rudrapatna lab for incredibly beneficial team meeting discussions.

**COMPETING INTERESTS**

VAR receives grant support from Merck Inc, Alnylam Pharmaceuticals Inc, Janssen Research and Development, and Genentech. AJB is a co-founder and consultant to Personalis and NuMedii; consultant to Mango Tree Corporation, and in the recent past, Samsung, 10x Genomics, Helix, Pathway Genomics, and Verinata (Illumina); has served on paid advisory panels or boards for Geisinger Health, Regenstrief Institute, Gerson Lehman Group, AlphaSights, Covance, Novartis, Genentech, and Merck, and Roche; is a shareholder in Personalis and NuMedii; is a minor shareholder in Apple, Meta (Facebook), Alphabet (Google), Microsoft, Amazon, Snap, 10x Genomics, Illumina, Regeneron, Sanofi, Pfizer, Royalty Pharma, Moderna, Sutro, Doximity, BioNtech, Invitae, Pacific Biosciences, Editas Medicine, Nuna Health, Assay Depot, and Vet24seven, and several other non-health related companies and mutual funds; and has received honoraria and travel reimbursement for invited talks from Johnson and Johnson, Roche, Genentech, Pfizer, Merck, Lilly, Takeda, Varian, Mars, Siemens, Optum, Abbott, Celgene, AstraZeneca, AbbVie, Westat, and many academic institutions, medical or disease specific foundations and associations, and health systems. AJB receives royalty payments through Stanford University, for several patents and other disclosures licensed to NuMedii and Personalis. AJB's research has been funded by NIH, Peraton (as the prime on an NIH contract), Genentech, Johnson and Johnson, FDA, Robert Wood Johnson Foundation, Leon Lowenstein Foundation, Intervalien Foundation, Priscilla Chan and Mark Zuckerberg, the Barbara and Gerson Bakar Foundation, and in the recent past, the March of Dimes, Juvenile Diabetes Research Foundation, California Governor's Office of Planning and Research, California Institute for Regenerative Medicine, L'Oreal, and Progenity. The authors have declared that no competing interests exist.

**FUNDING SOURCES**

DL and VAR were supported by a research grant from the Food and Drug Administration (FDA) of the U.S. Department of Health and Human Services (HHS). This grant was part of a financial assistance award [Center of Excellence in Regulatory Science and Innovation grant to the University of California, San Francisco (UCSF) and Stanford University, U01FD005978] totaling $79,250 with 100 percent funded by FDA/HHS. The contents are those of the author(s) and do not necessarily represent the official views of, nor an endorsement, by FDA/HHS, or the U.S. Government.

**SUPPLEMENTARY MATERIALS**

**Section S-A**: Different publicly available clinical inference tasks that we have evaluated our model on are described next:

1) **i2b2 2010 named entity recognition (NER)**[30]**:** This task addresses the fundamental question of concept recognition of types problems, treatments, and tests from clinical text. The dataset consists of sentences from 871 de-identified clinical reports annotated with these concepts. The task is set up as a named entity recognition task, where the goal is to identify the category of every token in text in terms of Begin-Inside-Out (BIO) tags for each of the 3 categories.
2) **i2b2 2010 relation classification (RE)**[30]**:** This task builds up on the i2b2-2010 concept recognition dataset to infer relations between pairs of concepts. These relations span the categories of problem-problem relations ('Problem indicates problem'), problem-test relations (test reveals problem, test conducted to investigate problem), and problem-treatment relations (treatment improves problem, treatment worsens problem, treatment causes problem, treatment administered for problem, treatment not administered because of problem). The task is set up as a classification task where the gold-standard concepts are known already, and the model needs to classify the correct relation between the given pairs of concepts.
3) **i2b2 2012 named entity recognition (NER)**[31]**:** Similar to the i2b2 2010 task for named entity recognition, this task focuses on recognizing clinical and temporal entities in text. The dataset consists of sentences from 310 de-identified clinical discharge summaries. The sentences are annotated with clinical concepts (problems, treatments, tests), departments, evidentials, and occurrences, in addition to temporal expressions that capture date, time, duration, and frequency.
4) **MedNLI**[32]**:** Given pairs of sentences, the task addresses the inference of entailment, contradiction, or neutral relations between these pairs. The dataset consists of nearly 14k sentence pairs that have been generated as a combination of sentences from the MIMIC-III corpus and manual input from domain experts to describe these relations.

**Section S-B**: Different publicly available biomedical and clinical language models we use are described next.

1) **BioBERT:** This is a BERT model pretrained on PubMed Central full-text articles and PubMed abstracts corpus. The model uses the same vocabulary as the BERT model pretrained on Wikipedia and BooksCorpus. We use version 1.1 of the model in our experiments.
2) **ClinicalBERT:** This is also a BERT model that has been trained on discharge summaries from the MIMIC-III dataset, using the BioBERT model as its starting point.
3) **SciBERT:** This is a BERT model trained on scientific articles, of which 82% is from the broad biomedical domain. We use the SciBERT model version that uses a custom vocabulary from these scientific articles.
4) **Biomed-RoBERTa:** This model is adapted from the RoBERTa-base model and has been continued to be trained on scientific articles from four domains, including the biomedical domain.
5) **BioLM:** This is also a RoBERTa model trained on a combination of PubMed articles and the MIMIC-III clinical corpus. We experiment with the base version that uses a custom vocabulary from PubMed articles and has been trained from scratch.
6) **GatorTron-OG:** This is one of the largest publicly available clinical language models, trained on a corpus of 82B words of de-identified clinical notes from the University of Florida Health System, 6.1B words from PubMed CC0, 2.5B words from WikiText, and 0.5B words from MIMIC-III. The model consists of a 50K token customized clinical vocabulary trained on the given data distribution. We use this model for evaluations instead of the contemporaneous GatorTron-S model because this model uses real-world clinical notes for pretraining instead of synthetic clinical notes.

**Table S1:** Note categories used for training the UCSF-BERT model and their frequencies in the corpus.

| Note Type | Frequency |
| --- | --- |
| Imaging | 20,144,949 |
| Progress Notes | 18,447,890 |
| Telephone Encounter | 9,078,213 |
| Consults | 3,623,887 |
| Note associated with ED Event | 3,104,966 |
| Pathology and Cytology | 3,050,597 |
| ECG | 2,937,844 |
| Assessment & Plan Note | 2,045,790 |
| History & Physical | 1,818,235 |
| Procedures | 1,164,351 |
| Operative Report | 1,104,313 |
| Problem Overview | 1,077,868 |
| Discharge Summary | 1,048,216 |
| ED Notes | 722,239 |
| ED Provider Notes | 705,203 |
| Echocardiography | 598,414 |
| Anesthesia Preprocedure Evaluation | 483,873 |
| UNDEFINED | 372,421 |
| Anesthesia Post-Op Note | 372,230 |
| Brief Op Note | 323,786 |
| Anesthesia Transfer of Care | 286,087 |
| GI | 218,155 |
| Ophthalmology | 202,457 |
| Consult | 202,261 |
| Imaging-Vasc | 173,082 |
| Cardiac Services | 170,941 |
| Interval H&P Note | 158,834 |
| IDG Notes | 153,468 |

| Note Type | Frequency |
|---|---|
| Imaging | 20,144,949 |
| Progress Notes | 18,447,890 |
| OR Anesthesia | 136,826 |
| Anesthesia Procedure Notes | 131,985 |
| PFT | 130,001 |
| Radiology study notes | 126,770 |
| imaging and procedure results | 118,627 |
| OR PostOp | 98,085 |
| OB | 89,455 |
| History & Physical (View-Only) | 83,251 |
| Significant Event | 77,417 |
| Anesthesia Postprocedure Evaluation | 77,414 |
| Cardiac Cath | 70,033 |
| Point-of-Care Imaging | 51,754 |
| Neurology | 48,471 |
| OR PreOp | 39,651 |
| Student Note | 38,449 |
| Electrophysiology | 33,096 |
| Face to Face | 24,319 |
| ECONSULT | 21,530 |
| L&D Delivery Note | 19,004 |
| OB Anesthesia Post-Partum Note | 16,236 |
| **Total** | **75,222,944** |

**Table S2:** 50 most frequent ICD-9 codes in the UCSF de-identified clinical data warehouse (2012 – early Mar 2021) that have been used for the ICD coding evaluation task.

| ICD-9 code | ICD-9 code description |
|---|---|
| 401.9 | Unspecified essential hypertension |
| 250.00 | Diabetes mellitus without mention of complication, type II or unspecified type, not stated as uncontrolled |
| V70.0 | Routine general medical examination at a health care facility |
| 272.4 | Other and unspecified hyperlipidemia |
| V20.2 | Routine infant or child health check |
| 185 | Malignant neoplasm of prostate |
| 174.9 | Malignant neoplasm of breast (female), unspecified |
| V42.0 | Kidney replaced by transplant |
| 427.31 | Atrial fibrillation |
| 696.1 | Other psoriasis |
| 729.5 | Pain in limb |
| 724.2 | Lumbago |
| 493.90 | Asthma,unspecified type, unspecified |
| 285.9 | Anemia, unspecified |
| V58.69 | Long-term (current) use of other medications |
| 244.9 | Unspecified acquired hypothyroidism |
| 719.46 | Pain in joint, lower leg |
| 401.1 | Benign essential hypertension |
| 530.81 | Esophageal reflux |
| 786.2 | Cough |
| 311 | Depressive disorder, not elsewhere classified |
| 300.00 | Anxiety state, unspecified |
| 789.00 | Abdominal pain, unspecified site |
| V76.2 | Screening for malignant neoplasms of cervix |
| V04.81 | Need for prophylactic vaccination and inoculation against influenza |

| ICD-9 code | ICD-9 code description |
|---|---|
| 401.9 | Unspecified essential hypertension |
| 250.00 | Diabetes mellitus without mention of complication, type II or unspecified type, not stated as uncontrolled |
| V70.0 | Routine general medical examination at a health care facility |
| 272.4 | Other and unspecified hyperlipidemia |
| V20.2 | Routine infant or child health check |
| 185 | Malignant neoplasm of prostate |
| V70.7 | Examination of participant in clinical trial |
| V45.89 | Other postprocedural status |
| 786.50 | Chest pain, unspecified |
| 780.60 | Fever, unspecified |
| 780.79 | Other malaise and fatigue |
| 338.29 | Other chronic pain |
| 272.0 | Pure hypercholesterolemia |
| V49.89 | Other specified conditions influencing health status |
| 628.9 | Infertility, female, of unspecified origin |
| 599.0 | Urinary tract infection, site not specified |
| 278.00 | Obesity, unspecified |
| 724.5 | Backache, unspecified |
| V22.1 | Supervision of other normal pregnancy |
| 692.9 | Contact dermatitis and other eczema, unspecified cause |
| V76.12 | Other screening mammogram |
| 199.1 | Other malignant neoplasm without specification of site |
| 784.0 | Headache |
| 203.00 | Multiple myeloma, without mention of having achieved remission |
| V72.31 | Routine gynecological examination |

| ICD-9 code | ICD-9 code description |
|---|---|
| 401.9 | Unspecified essential hypertension |
| 250.00 | Diabetes mellitus without mention of complication, type II or unspecified type, not stated as uncontrolled |
| V70.0 | Routine general medical examination at a health care facility |
| 272.4 | Other and unspecified hyperlipidemia |
| V20.2 | Routine infant or child health check |
| 185 | Malignant neoplasm of prostate |
| 465.9 | Acute upper respiratory infections of unspecified site |
| 272.2 | Mixed hyperlipidemia |
| 786.09 | Other respiratory abnormalities |
| 723.1 | Cervicalgia |
| 477.9 | Allergic rhinitis, cause unspecified |
| 564.00 | Constipation, unspecified |

**Table S3:** 50 medication therapeutic classes coded in the deidentified clinical data warehouse at UCSF

| |
|---|
| Miscellaneous Medical Supplies, Devices, Non-Drug |
| Unclassified Drug Products |
| Skin Preps |
| Cough/Cold Preparations |
| Gastrointestinal |
| Elect/Caloric/H2O |
| Vitamins |
| Analgesics |
| Antibiotics |
| EENT Preps |
| Biologicals |
| Antineoplastics Hormones |
| Anesthetics |
| Cardiovascular |
| Diagnostic |
| Pre-Natal Vitamins |
| Antihistamine And Decongestant Combination |
| Psychotherapeutic Drugs |
| Herbals |
| Antiarthritics |
| Antihistamines |
| Cardiac Drugs |
| Antiasthmatics |
| CNS Drugs |
| Blood |
| Antifungals |
| Contraceptives |
| Antihyperglycemics |
| Autonomic Drugs |
| Antivirals |
| Sedative/Hypnotics |
| Diuretics |
| Antiinfectives/Miscellaneous |
| Anticoagulants |
| Immunosuppressants |
| Thyroid Preps |
| Anti-Obesity Drugs |
| Antiparkinson Drugs |
| Analgesic And Antihistamine Combination |
| Colony Stimulating Factors |
| Muscle Relaxants |
| Antiparasitics |
| Antiplatelet Drugs |
| Antiinflam.Tumor Necrosis Factor Inhibiting Agents |
| Smoking Deterrents |
| Antidotes |
| Anti-infectives |
| Miscellaneous Medical Supplies Or Devices |
| Antiallergy |

**Table S4:** Size and length of datasets for public and UCSF-specific evaluation tasks

| Dataset description | | Dataset size | | | Length of dataset in words | | | |
|---|---|---|---|---|---|---|---|---|
| Dataset | Task category | n_train | n_dev | n_test | Min | Max | Median | Mean |
| **I2B2-2010-NER** | NER | 14354 | 1961 | 27625 | 1 | 180 | 5 | 8.1 |
| **I2B2-2010-RE** | Classification | 21384 | 872 | 43000 | 2 | 204 | 28 | 35.5 |
| **I2B2-2012-NER** | NER | 6615 | 831 | 5665 | 1 | 87 | 11 | 12.1 |
| **MedNLI** | Classification | 11232 | 1395 | 1422 | 4 | 148 | 19 | 21.4 |
| **ICD coding (top 50)** | Classification | 22728 | 2841 | 2821 | 223 | 12859 | 999 | 1205.4 |
| **Therapeutic class prediction** | Classification | 17769 | 2150 | 2214 | 233 | 13241 | 1319 | 1564.6 |

**Table S5:** Mean and median sequence length of the dataset when encoded with different vocabularies. The % diff values reflect the percentage differences in the sequence lengths compared to encoding the dataset with BERT vocabulary.

| Task | Model | Vocabulary | Mean length | % diff mean | Median length | % diff median |
|---|---|---|---|---|---|---|
| ICD9 top50 coding | BERT, clinical BERT, BioBERT | Wikipedia, BookCorpus | 2465 | Base | 2032 | Base |
| | SciBERT | Scientific articles | 2252 | -9% | 1846 | -9% |
| | BioLM | PubMed | 2172 | -12% | 1833 | -10% |
| | UCSF-BERT | UCSF clinical notes | **1945** | **-21%** | **1590** | **-22%** |
| Therapeutic class inference | BERT, clinical BERT, BioBERT | Wikipedia, BookCorpus | 3218 | Base | 2744 | Base |
| | SciBERT | Scientific articles | 2990 | -7% | 2550 | -7% |
| | BioLM | PubMed | 2774 | -14% | 2380 | -13% |
| | UCSF-BERT | UCSF clinical notes | **2591** | **-20%** | **2196** | **-20%** |
| MedNLI | BERT, clinical BERT, BioBERT | Wikipedia, BookCorpus | 39 | Base | 33 | Base |
| | SciBERT | Scientific articles | 34 | -13% | 29 | -12% |
| | BioLM | PubMed | 31 | -21% | 27 | -18% |
| | UCSF-BERT | UCSF clinical notes | **31** | **-21%** | **26** | **-21%** |
| i2b2 2010 RE | BERT, clinical BERT, BioBERT | Wikipedia, BookCorpus | 65 | Base | 47 | Base |
| | SciBERT | Scientific articles | 59 | -9% | 42 | -11% |
| | BioLM | PubMed | 55 | -15% | 39 | -17% |
| | UCSF-BERT | UCSF clinical notes | **52** | **-20%** | **38** | **-19%** |
| i2b2 2012 NER | BERT, clinical BERT, BioBERT | Wikipedia, BookCorpus | 21 | Base | 18 | Base |
| | SciBERT | Scientific articles | 19 | -10% | 17 | -6% |
| | BioLM | PubMed | **17** | **-19%** | 15 | -17% |
| | UCSF-BERT | UCSF clinical notes | **17** | **-19%** | 15 | -17% |

**Table S6**: Randomly sampled errors made by the UCSF-BERT model on the i2b2 2010 named entity recognition task for recognizing problems, treatments and tests as concepts. Bold reflects the error, underline is for the correctly recognized surrounding concept, if relevant.

| Error Category | Example |
|---|---|
| Predicted as treatment but not annotated as such | The patient was a 53-year-old male with a longstanding history of renal disease with multiple complications and problems related to **vascular access** and hypercoagulable state. |

|  | The patient had undergone two abdominal procedures following **explantation** .<br>*(Potentially incorrect annotation)* |
|---|---|
|  | Baby was vigorous at birth and received **routine DR Johnson** .<br>*(Contextually correct, but unlear semantics of tokens)* |
|  | They have opted for **conservative management** given the size .<br>*(Potentially incorrect annotation)* |
|  | The patient 's electrolytes and **fluids** were monitored closely during his admission . |
| Not predicted as treatment | For this reason , he was taken again to the operating room on 01/08/98 , at which time a gastrojejunostomy was performed with stapling **across the gastroduodenal junction** , introduction of five Jackson-Pratt drains into the retroperitoneal mass , cholecystotomy tube , gastrostomy tube , and jejunostomy tube .<br>*(Potentially incorrect annotation)* |
|  | He had **daily transfusions** of FFP and blood , and continued coagulopathy . |
|  | During the seizure , he bit his tongue , resulting in a large bleed from his tongue , which was **sutured** by the ENT Service , however , his continued coagulopathy resulted in bleeding from his nasopharynx , which could never be identified , and also upper gastrointestinal bleeding . |
|  | He was transferred to the newborn nursery for **further care** . |
|  | At this point CMED was consulted again and he was transferred to the CMED triage for **further care** . |
|  | Albuterol nebulizers 2.5 mg q.4h. and Atrovent nebulizers 0.5 mg q.4h. , please alternate albuterol and Atrovent ; Rocaltrol 0.25 mcg per **NG tube** q.d.; calcium carbonate 1250 mg per **NG tube** q.i.d.; vitamin B12 1000 mcg IM q. month , next dose is due Nov 18 ; diltiazem 60 mg per **NG tube** t.i.d.; ferrous sulfate 300 mg per **NG** t.i.d.; Haldol 5 mg IV q.h.s.; hydralazine 10 mg IV q.6h. p.r.n. hypertension ; lisinopril 10 mg per **NG tube** q.d.; Ativan 1 mg per **NG tube** q.h.s.; Lopressor 25 mg per **NG tube** t.i.d.; Zantac 150 mg per **NG tube** b.i.d.; multivitamin 10 ml per **NG tube** q.d.; Macrodantin 100 mg per **NG tube** q.i.d. x 10 days beginning on 11/3/00 . |
|  | I stress that she cannot be **anticoagulated** . |
|  | Mr. Liley was discharged in stable condition , ambulating and voiding independently , and with **adequate pain control** . |
|  | He was advanced to **clears** which he tolerated and was restarted on a low residue diet . |
|  | After catheter placement , the patient was started on **sips** . |
|  | **MEDS** |
| Predicted as treatment instead of problem | The patient was found to have **substantial auto PEEP** with an elevated dead space. |
| | PULM - CTAB no w / r / r ABD -nt / nd ; **incision** c / d / I ; discrete 3x3 cm slightly firm protuberant mass to the R of umbilicus ; no rebound , no guarding . |
| Predicted as test instead of problem | Stool cultures have been sent numerous times as well as analysis for **C. difficile toxin** . *(Potentially incorrect annotation)* |
| | He had 3 specimens sent for **C. difficile** which were negative . |
| Predicted as treatment instead of test | Significant for motor vehicle accident in 1966 , requiring **exploratory laparotomy** and splenectomy , tonsillectomy as a child , pneumonia in the past , hepatitis C and recent diagnosis of hepatocellular carcinoma . |
| | Right arm was 180/80 , left 188/72 , and **A-line** was 151/61 . |
| Predicted as test instead of treatment | **Tidal volume** is 700 with a rate of 10 , 10 of pressure support , and 10 of PEEP , and 100% FiO2 , satting 100% . |
| Predicted as test but not annotated as such | **The physiology** at the time of transfer to the Medical Intensive Care Unit was consistent with a septic process . |
| | The patient had recently prior to this admission , had a rise in his **alpha feta protein** and a biopsy of his liver had shown hepatocellular carcinoma . |
| | Significant for normal **head , eyes , ears , nose , throat exam** with no scleral icterus . |
| | Severe hypertension , angina , GERD , peptic ulcer disease , status post Billroth II , ovarian cancer , status post TAH-BSO in 1987 with a negative **second look laparotomy** in 1988 , chronic anemia , chronic left ear infection . |
| | **PULSE** : |
| | **The patient 's electrolytes** and fluids were monitored closely during his admission . |
| Not predicted as test | He had **3 specimens** sent for C. difficile which were negative . *(Potentially incorrect annotation)* |
| Predicted as problem but not annotated as such | Accompanying this during this period were elevations **in** liver **function studies** and amylase and lipase consistent with pancreatic inflammation . |
| | Extensive discussions were carried out with Mrs. **Less condition** was irretrievable . |
| | His immunosuppression was OKT3 and Solu Medrol , and he underwent hemodialysis and ultra filtration to remove **fluid** . |
| | Last week , she had copious amounts of stool in the range of 500 cc to 1000 cc of **liquid stool** per day . |
| | Given **these findings** , she underwent a Sestamibi scan . |
| | The operative report shows that there was an abnormal area below the trachea to the left which represented **a cavity** between the esophagus , trachea , aorta , |

|  | and spine that was full of clot . |
|---|---|
|  | CT scan showed an anterior mediastinal mass compressing the trachea with **air** . |
|  | However , it did show an anterior mediastinal mass compressing the trachea with **a density** consistent with a hematoma . |
|  | Given **these findings** , she underwent a CT scan which showed a retrotracheal mass that appeared to obliterate the esophagus . |
|  | His bowel function resumed and in fact he had **loose stool** . |
|  | No **active issues** . |
| Not predicted as problem | He remained pressor dependent , as well as **dependent** upon mechanical ventilatory assistance . |
|  | He had **an ongoing requirement** for high volume resuscitation and was persistently acidotic despite CV-VVH in the face of bicarbonate replenishment . |
|  | The patient also appeared to have stool from **his Jackson-Pratt site** and right oblique incision . |
|  | Mr. Villesatelkscurb is a fifty year old black gentleman with a long history of hepatitis C , believed to be contracted during transfusions in **a previous trauma** . |
|  | **Large for gestational age appearance** of an infant with tachypnea , grunting and flaring . |
|  | Stable , with **active problems** being respiratory failure , urinary tract infection , and bleeding diathesis . *(Potential annotation issue: generic term)* |
|  | She had no **problems** with this . *(Potential annotation issue: generic term)* |
|  | Her tracheostomy is functioning well , and we have had no **problems** with it . *(Potential annotation issue: generic term)* |
|  | Blood cultures from 04/07 showed **NGTD** . |
|  | PULM - CTAB no w / r / r ABD **-nt** / nd ; incision c / d / I ; discrete 3x3 cm slightly firm protuberant mass to the R of umbilicus ; no rebound , no guarding . |
| Similar concepts get integrated with intermediate words | <u>Bloody drainage</u> **from** <u>his increasingly distended abdomen</u> was noted . |
|  | Initially , the patient did well in the Intensive Care Unit , mentally alert , and oriented , however , he continued to be coagulopathic and was requiring **large amounts** <u>of blood</u> **and** <u>FFP transfusions</u> , with a resultant pulmonary edema . |
| | - Seek medical attention for fevers ( temp > 101.5 ) , worsening pain , drainage or <u>excessive bleeding</u> **from** <u>incision</u> , chest pain , shortness of breath , or <u>any other symptoms</u> **of concern** . |
| Modifiers included with concept | It was felt that the patient had **continued** <u>massive retroperitoneal contamination</u> and probable multiple enteric fistulae from an ischemic bowel . |
| | **On admission** <u>medications</u> included Lasix and iron . |
| | However , he became **progressively more** <u>mentally obtunded</u> . |
| | Her hematocrit continued to drop over the next couple of days despite normal platelet count , normal coags , and no obvious **source of** <u>bleeding</u> . |
| | She initially presented to Brottlake Medical Center on 9/15/00 complaining of <u>acute onset dysphasia</u> **to liquids and solids** . |
| Modifiers not included in concept | While in the ICU he developed seizures , requiring **large amounts of** <u>intravenous Valium</u> to break his seizures , and he was maintained on Tegretol and Dilantin . |
| | He was noted to be <u>large</u> **for gestational age** with a weight of 4525 grams . |
| | We have given **her** <u>Lovenox</u> in the past , and she continues to bleed . |
| | She underwent a parathyroid exploration with resection of a ruptured ectopic right upper parathyroid adenoma , <u>a biopsy</u> **of a normal left and right lower parathyroids** , resection of a left upper parathyroid , and tracheostomy . |
| | The operative report shows that there was <u>an abnormal area below</u> **the trachea** to the left which represented a cavity between the esophagus , trachea , aorta , and spine that was full of clot . |
| | Her anterior neck and chest had **a 1** <u>cm to 2 cm area of ecchymosis</u> . |
| | PULM - CTAB no w / r / r ABD -nt / nd ; incision c / d / I ; discrete 3x3 cm slightly firm <u>protuberant mass</u> **to the R of umbilicus** ; no rebound , no guarding . |
| Predicted as a different IOB tag but right type | Endoscopy never revealed specific sources of **bleeding** , although he appeared to have a diffuse duodenitis . |
| | 2) 2+ **E.** coli sensitive to Ampicillin and Levofloxacin , |
| Only subconcept recognized | She had <u>an anastomosis</u> **between the** <u>left internal mammary</u> **artery and the left anterior descending artery** , saphenous vein graft to RCA , saphenous vein graft to OM . |
| | It was believed that she was continuing to **bleed** <u>in her chest</u> . |

**Table S7:** Manually curated clinical language inference examples to assess numerical, temporal and clinical reasoning skills of the UCSF-BERT model. *Glucose* values were considered as *normal* within the range 70–100; *Blood pressure (BP)* was considered as *normal* until the upper limit of 120; BP should be treated above systolic 130

vs 140 vs 150 (elderly) (controversial); *serum albumin* was considered as *normal* within the range 3.5–5.5 g/dL (35-55 g/L); *Globulins, total* was considered as *normal* within the range 2.5–3.5 g/dL (25–35 g/L); *Epinephrine, plasma (supine)* was considered as *normal* when less than 75 ng/L; *Body Mass Index (BMI)* within the range 18.5–24.9 was considered *normal*, within 25–29.9 range was considered as *overweight*, and ≥30 was considered to be *obese*; *serum calcium* was considered as *normal* within the range 9–10.5 mg/dl; *low-density lipoprotein (LDL)* was considered *desirable* when less than or equal to 130 mg/dL (3.36 mmol/L); *Potassium, serum* was considered as *normal* in the range 3.5–5.0 meq/L (3.5-5.0 mmol/L)

| Premise | Hypothesis | Label |
| --- | --- | --- |
| The patient has chest pain and ST segment elevation on ECG | The patient may be having an MI | Entailment |
| The patient's blood glucose is 600 | The patient is having hyperglycemia | Entailment |
| The patient's blood glucose is 600 | The patient is having hypoglycemia | Contradiction |
| The patient's blood glucose is 30 | The patient is having hypoglycemia | Entailment |
| The patient took his medication and then the patient got sick | The patient got sick after taking medication | Entailment |
| The patient took his medication and then the patient got sick | The patient got sick before taking medication | Contradiction |
| The patient took his medication and then later the patient got sick | The patient got sick before taking medication | Contradiction |
| First came the drug and then came the reaction | The reaction came before the drug | Contradiction |
| Blood glucose is 800 | The patient has hypoglycemia | Contradiction |
| Blood glucose is 800 | The patient has hyperglycemia | Entailment |
| Blood glucose is 40 | The patient has hyperglycemia | Contradiction |
| Blood glucose is 25 | The patient has hyperglycemia | Contradiction |
| Blood glucose is 25 | The patient has hypoglycemia | Entailment |
| Blood glucose is 25 | The patient has low blood glucose | Entailment |
| The patient has glucose of 25 | The patient has low blood glucose | Entailment |
| The patient has glucose of 10 | The patient has low blood glucose | Entailment |
| The patient has glucose of 500 | The patient has high blood glucose | Entailment |
| The patient has glucose of 25 | The patient has high blood glucose | Contradiction |
| The patient has glucose of 25 | The patient has low blood glucose | Entailment |
| The patient has glucose of 25 | The patient has hypoglycemia | Entailment |
| The patient has glucose of 25 | The patient has low blood sugar | Entailment |
| The patient has blood glucose of 25 | The patient has low blood sugar | Entailment |
| The patient's blood glucose is 25 | The patient has low blood sugar | Entailment |
| The patient's blood glucose is 25 | The patient has hypoglycemia | Entailment |
| The patient's blood glucose is 25 | The patient has hyperglycemia | Contradiction |
| The patient's blood pressure is 180/110 | The patient has hypertension | Entailment |

| Premise | Hypothesis | Label |
|---|---|---|
| The patient's blood pressure is 180/110 | The patient has hypotension | Contradiction |
| The patient's blood pressure is 60/0 | The patient has hypotension | Entailment |
| The patient's blood pressure is 60/0 | The patient is in shock | Entailment |
| The patient's blood pressure is 60/0 | The patient may be in shock | Entailment |
| The patient can not smell coffee | The patient has anosmia | Entailment |
| The patient can not smell anything | The patient has anosmia | Entailment |
| The patient can not hear anything | The patient may be deaf | Entailment |
| The patient has septicemia | The patient may have a cytokine storm | Entailment |
| The patient has septicemia | The patient has bacteria growing in the blood stream | Entailment |
| The patient has a history of hypertension. Now the patient is short of breath with any exertion | The patient may have CHF | Entailment |
| The patient has a history of hypertension. Now the patient is short of breath with any exertion and a reduced left ventricular ejection fraction | The patient may have CHF | Entailment |
| The patient has a BMI of 20 | The patient is obese | Contradiction |
| The patient has a BMI of 35 | The patient is obese | Entailment |
| The patient has a BMI of 60 | The patient is obese | Entailment |
| The patient has a BMI of 25 | The patient has a healthy weight | Contradiction |
| The patient has a BMI of 25 | The patient is overweight | Entailment |
| The patient's pulse is 120 | The patient has tachycardia | Entailment |
| The patient's pulse is 40 | The patient has bradycardia | Entailment |
| The patient's pulse is 120 | The patient has bradycardia | Contradiction |
| The patient's has right lower quadrant pain and a high fever | The patient may have a ruptured appendix | Entailment |
| The patient's has right lower quadrant pain and a high fever | The patient may have appendicitis | Entailment |
| The patient's had the sudden onset of pain and tenderness on the right abdomen. | The patient may have appendicitis | Entailment |
| The patient had the sudden onset of pain and tenderness on the right lower abdomen. | The patient may have appendicitis | Entailment |
| The patient had the sudden onset of pain and tenderness on the right lower abdomen. The abdomen shows board-like rigidity | The patient may have appendicitis | Entailment |
| The patient had the sudden onset of pain and tenderness on the right lower abdomen. The abdomen shows board-like rigidity. The patient has a low grade fever. | The patient may have appendicitis | Entailment |
| The patient had the sudden onset of pain and tenderness on the right lower abdomen. The abdomen shows board-like rigidity. The patient has a low grade fever. | The patient may have an acute abdomen | Entailment |

| Premise | Hypothesis | Label |
|---|---|---|
| The patient had the sudden onset of pain and tenderness on the right lower abdomen. The abdomen shows board-like rigidity. The patient has a low grade fever. | The patient should be evaluated by a surgeon | Entailment |
| The patient has positive blood cultures for e. coli | The patient has bacteremia | Entailment |
| The patient has positive blood cultures for e. coli | The patient has septicemia | Neutral |
| The patient is taking a blood pressure medicine. The patient gets dizzy when he suddenly stands up. | The patient may be having an adverse drug reaction to the blood pressure medicine | Entailment |
| The patient is taking a blood pressure medicine. The patient is hungry. | The patient may be having an adverse drug reaction to the blood pressure medicine | Neutral |
| The patient is taking a blood pressure medicine. The patient is in his usual state of good health. | The patient may be having an adverse drug reaction to the blood pressure medicine | Contradiction |
| The patient is taking an immune suppressant medication. The patient has an opportunistic infection | The patient may be having an adverse drug reaction to the medication | Entailment |
| The patient is taking an immune suppressant medication. The patient has an infection with an opportunistic organism | The patient may be immune compromised | Entailment |
| The patient has suicidal ideation and plans to kill himself | The patient should have a consultation with a psychiatrist | Entailment |
| The patient has suicidal ideation and plans to kill himself | The patient is at risk for suicide | Entailment |
| The patient denies suicidal ideation | The patient is at risk for suicide | Contradiction |
| The patient is happy | The patient is at risk for suicide | Contradiction |
| The patient is an adolescent, and began taking an SSRI 1 week ago | The patient is at risk for suicide | Entailment |
| The patient is an adolescent, and began taking an SSRI 20 weeks ago | The patient is at risk for suicide | Neutral |
| The patient is an adolescent | The patient is at risk for suicide | Neutral |
| The patient is 55 years old, and has crushing, substernal chest pain | R/O Myocardial Infarction | Entailment |
| The patient wants a routine physical examination | R/O Myocardial Infarction | Neutral |
| Blood glucose is 23 | The patient has low blood glucose | Entailment |
| Blood glucose is 23 | The patient has hypoglycemia | Entailment |
| Blood glucose is 23 | The patient has hyperglycemia | Contradiction |
| Blood glucose is 27 | The patient has hypoglycemia | Entailment |
| Blood glucose is 47 | The patient has hypoglycemia | Entailment |
| Blood glucose is 57 | The patient has hypoglycemia | Entailment |
| Blood glucose is 77 | The patient has hypoglycemia | Contradiction |
| Blood glucose is 70 | The patient has hypoglycemia | Contradiction |
| Blood glucose is 69 | The patient has hypoglycemia | Entailment |
| Blood glucose is 61 | The patient has hypoglycemia | Entailment |
| Blood glucose is 59 | The patient has hypoglycemia | Entailment |
| Blood glucose is 64 | The patient has hypoglycemia | Entailment |
| Blood glucose is 60 | The patient has hypoglycemia | Entailment |
| Blood glucose is 62 | The patient has hypoglycemia | Entailment |
| Blood glucose is 61 | The patient has hypoglycemia | Entailment |
| The patient's blood pressure is 160/80 | The patient has hypertension | Entailment |
| The patient's blood pressure is 155/80 | The patient has hypertension | Entailment |
| The patient's blood pressure is 150/80 | The patient has hypertension | Entailment |
| The patient's blood pressure is 140/80 | The patient has hypertension | Entailment |
| The patient's blood pressure is 130/80 | The patient has hypertension | Entailment |
| The patient's blood pressure is 135/80 | The patient has hypertension | Entailment |
| The patient's blood pressure is 137/80 | The patient has hypertension | Entailment |
| The patient's blood pressure is 136/80 | The patient has hypertension | Entailment |
| The patient's blood pressure is 135/80 | The patient has hypertension | Entailment |
| The patient's blood pressure is 134/80 | The patient has hypertension | Entailment |
| The patient's blood pressure is 133/80 | The patient has hypertension | Entailment |
| The patient's blood pressure is 132/80 | The patient has hypertension | Entailment |
| The patient's blood pressure is 131/80 | The patient has hypertension | Entailment |
| The patient's blood pressure is 130/80 | The patient has hypertension | Entailment |
| The patient's blood pressure is 133/80 | The patient has hypertension | Entailment |
| The patient's blood pressure is 136/80 | The patient has hypertension | Entailment |
| The patient's blood pressure is 137/80 | The patient has hypertension | Entailment |
| The patient's blood pressure is 138/80 | The patient has hypertension | Entailment |
| The patient's blood pressure is 139/80 | The patient has hypertension | Entailment |
| The patient's blood pressure is 140/80 | The patient has hypertension | Entailment |
| The patient's blood pressure is 141/80 | The patient has hypertension | Entailment |
| The patient's blood pressure is 142/80 | The patient has hypertension | Entailment |
| The patient's blood pressure is 143/80 | The patient has hypertension | Entailment |
| The patient's blood pressure is 144/80 | The patient has hypertension | Entailment |
| The patient's blood pressure is 145/80 | The patient has hypertension | Entailment |
| The patient's blood pressure is 146/80 | The patient has hypertension | Entailment |
| The patient's blood pressure is 147/80 | The patient has hypertension | Entailment |
| The patient's blood pressure is 148/80 | The patient has hypertension | Entailment |
| The patient's blood pressure is 149/80 | The patient has hypertension | Entailment |
| The patient's blood pressure is 150/80 | The patient has hypertension | Entailment |
| The patient's blood pressure is 160/80 | The patient has hypertension | Entailment |
| The patient's blood pressure is 170/80 | The patient has hypertension | Entailment |

| Premise | Hypothesis | Label |
| --- | --- | --- |
| The patient's blood pressure is 180/80 | The patient has hypertension | Entailment |
| The patient's blood pressure is 190/80 | The patient has hypertension | Entailment |
| The patient's blood pressure is 200/80 | The patient has hypertension | Entailment |
| The patient's blood pressure is 210/80 | The patient has hypertension | Entailment |
| The patient's blood pressure is 220/80 | The patient has hypertension | Entailment |
| The patient's blood pressure is 230/80 | The patient has hypertension | Entailment |
| The patient's serum calcium is 7 | The patient has hypocalcemia | Entailment |
| The patient's serum calcium is 6 | The patient has hypocalcemia | Entailment |
| The patient's serum calcium is 8 | The patient has hypocalcemia | Entailment |
| The patient's serum calcium is 9 | The patient has hypocalcemia | Contradiction |
| The patient's serum calcium is 9 | The patient has hypercalcemia | Contradiction |
| The patient's serum calcium is 9 | The patient has normal serum calcium | Entailment |
| The patient's serum calcium is 10 | The patient has normal serum calcium | Entailment |
| The patient's serum calcium is 11 | The patient has normal serum calcium | Contradiction |
| The patient's serum calcium is 12 | The patient has normal serum calcium | Contradiction |
| The patient's serum calcium is 13 | The patient has normal serum calcium | Contradiction |
| The patient's serum calcium is 13 | The patient has hypercalcemia | Entailment |
| The patient's serum calcium is 13.0 | The patient has hypercalcemia | Entailment |
| The patient's serum calcium is 14 | The patient has hypercalcemia | Entailment |
| The patient's serum calcium is 15 | The patient has hypercalcemia | Entailment |
| The patient's serum calcium is 15.1 | The patient has hypercalcemia | Entailment |
| The patient's serum calcium is 16 | The patient has hypercalcemia | Entailment |
| The patient's serum calcium is 16.5 | The patient has hypercalcemia | Entailment |
| The patient's serum calcium is 17 | The patient has hypercalcemia | Entailment |
| The patient's serum calcium is 17.0 | The patient has hypercalcemia | Entailment |
| The patient's serum calcium is 17.5 | The patient has hypercalcemia | Entailment |
| The patient's serum albumin is 2.5 g/dL | The patient has low albumin | Entailment |
| The patient's serum albumin is 4 g/dL | The patient has low albumin | Contradiction |
| The patient's serum albumin is 5 g/dL | The patient has low albumin | Contradiction |
| The patient's serum albumin is 10 g/dL | The patient has low albumin | Contradiction |
| The patient's serum albumin is 50 g/dL | The patient has low albumin | Contradiction |
| The patient's serum albumin is 100 g/dL | The patient has low albumin | Contradiction |
| The patient's serum albumin is 100 g/dL | The patient has hypoalbuminemia | Contradiction |
| The patient's serum albumin is 100 g/dL | The patient has hyperalbuminemia | Entailment |
| The patient's serum globulins is 3.0 | The patient has low globulins | Contradiction |
| The patient's LDL is 200 | The patient has high LDL | Entailment |
| The patient's LDL is 100 | The patient has high LDL | Contradiction |

| Premise | Hypothesis | Label |
|---|---|---|
| The patient's LDL is 150 | The patient has high LDL | Entailment |
| The patient's LDL is 140 | The patient has high LDL | Entailment |
| The patient's plasma epinephrine is 50 ng/L | The patient's plasma epinephrine is high | Contradiction |
| The patient's plasma epinephrine is 100 ng/L | The patient's plasma epinephrine is high | Entailment |
| The patient's plasma epinephrine is 200 ng/L | The patient's plasma epinephrine is high | Entailment |
| The patient's serum potassium is 2.0 | The patient has hypokalemia | Entailment |
| The patient's serum potassium is 6.0 | The patient has hypokalemia | Contradiction |
| The patient's serum potassium is 6.0 | The patient has hyperkalemia | Entailment |
| The patient's serum potassium is 6.0 | The patient has hyponatremia | Neutral |
| The patient's serum potassium is 6.0 | The patient has hypernatremia | Neutral |
| The patient's serum potassium is 6.0 | The patient has high cholesterol | Neutral |
| The patient's hematocrit is 30 | The patient is anemic | Entailment |
| The patient's hematocrit is 50 | The patient is anemic | Contradiction |